\documentclass[runningheads]{llncs}

\usepackage{eccv}

\usepackage{eccvabbrv}

\usepackage{graphicx}

\usepackage[accsupp]{axessibility}  

\usepackage{url}
\usepackage{graphicx}    
\usepackage{wrapfig}
\usepackage[font=small]{caption}
\usepackage{subcaption}
\usepackage{amsmath,amsfonts,bm}
\usepackage{colortbl}
\newcommand{\ourmethod}{\textit{UniMedI}}

\definecolor{graycolor}{gray}{.895} 

\usepackage{hyperref}

\usepackage{orcidlink}

\usepackage{booktabs}

\begin{document}

\title{Unified Medical Image Pre-training in Language-Guided Common Semantic Space} 

\titlerunning{Unified Medical Image Pre-training.}

\author{Xiaoxuan He$^{2}$\thanks{Work done during an internship with Microsoft Research.}, Yifan Yang$^{1}$, Xinyang Jiang$^{1}$, Xufang Luo$^{1}$\thanks{Correspondence to Xufang Luo, contact: \href{mailto:xufluo@microsoft.com}{xufluo@microsoft.com}}, Haoji Hu$^{2}$, Siyun Zhao$^{1}$, Dongsheng Li$^{1}$, Yuqing Yang$^{1}$, Lili Qiu$^{1}$\\
} 
\authorrunning{He et al.}

\institute{Microsoft Research \and
Zhejiang University}

\maketitle

\begin{abstract}
    Vision-Language Pre-training (VLP) has shown the merits of analysing medical images. It efficiently learns visual representations by leveraging supervisions in their corresponding reports, and in turn facilitates analysis and interpretation of intricate imaging data. 
    However, such observation is predominantly justified on single-modality data (mostly 2D images like X-rays), adapting VLP to learning unified representations for medical images in real scenario remains an open challenge. This arises from medical images often encompass a variety of modalities, especially modalities with different dimensions (e.g., 3D images like Computed Tomography), and there are almost no paired multi-dimension data here.
    To overcome the aforementioned challenges, we propose an \textbf{U}nified \textbf{Med}ical \textbf{I}mage Pre-training framework, namely \ourmethod{}, which utilizes diagnostic reports as common semantic space to create unified representations for diverse modalities of medical images (especially for 2D and 3D images). 
    Under the text's guidance, \ourmethod{} effectively select text-related 2D slices from sophisticated 3D volume, which acts as pseudo-pairs to bridge 2D and 3D data, ultimately enhancing the consistency across various medical imaging modalities. 
    To demonstrate the effectiveness and versatility of \ourmethod{}, we evaluate its performance on both 2D and 3D images across several different datasets, covering a wide range of medical image tasks such as classification, segmentation, and retrieval. \ourmethod{} has demonstrated superior performance in downstream tasks, showcasing its effectiveness in establishing a universal medical visual representation.
    \keywords{Vision-Language Pre-Training \and Medical Image \and Multi-Modality}
\end{abstract}

\section{Introduction}

In recent years, deep models with Visual-Language Pre-training (VLP) \cite{huang2021gloria, boecking2022making,bannur2023learning} attract lots of attention in medical image analysis, as VLP reduces the need for costly and time-consuming manual annotations by leveraging the vast amount of information in radiology reports and unlabeled data.
Despite these success, further expanding the data scale for medical VLP remains non-trivial, because the availability of single-modality medical images is limited, especially when compared to the general domain. This introduces a strong need to integrate multi-modality medical images (e.g., X-rays, Computed Tomography (CT) and Magnetic Resonance Imaging (MRI)) within a unified Vision-Language (VL) framework. However, fully leveraging the information across multi-modal images within this VL framework is unexplored.

\begin{figure}[t]
\centering
    \begin{subfigure}[b]{0.48\textwidth}   
        \centering  
        \includegraphics[width=\textwidth]{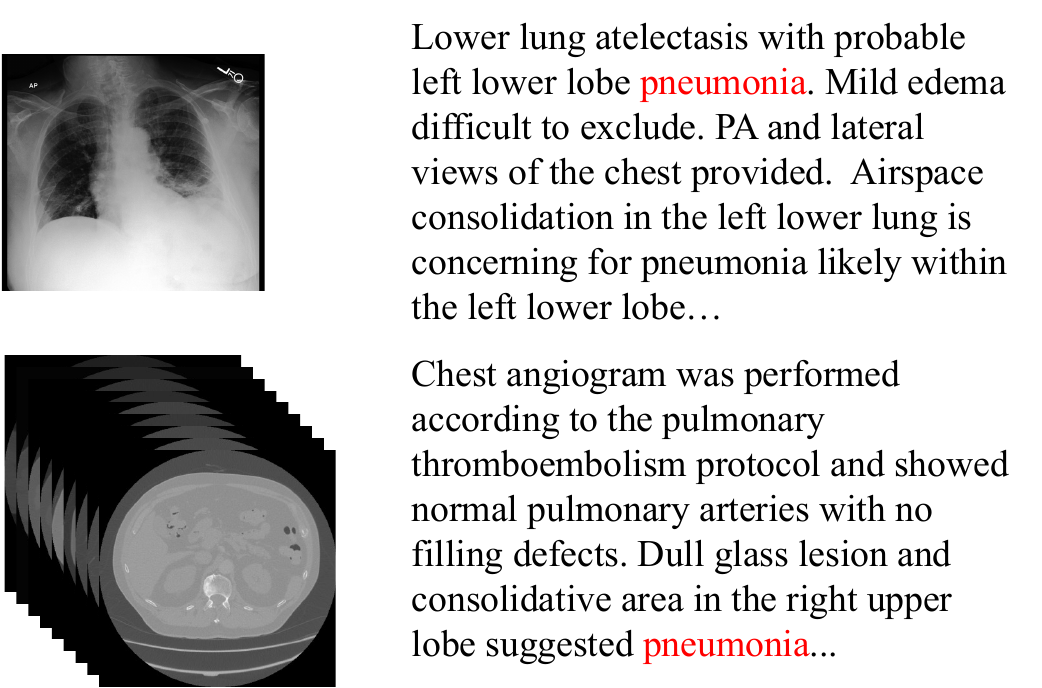}
        \caption{An example showing X-ray (up) and CT scan (down) both demonstrate similar abnormality, recording in the reports (right).}  
        \label{fig:example}
    \end{subfigure}\hfill
    \begin{subfigure}[b]{0.48\textwidth}   
        \centering  
        \includegraphics[width=\textwidth]{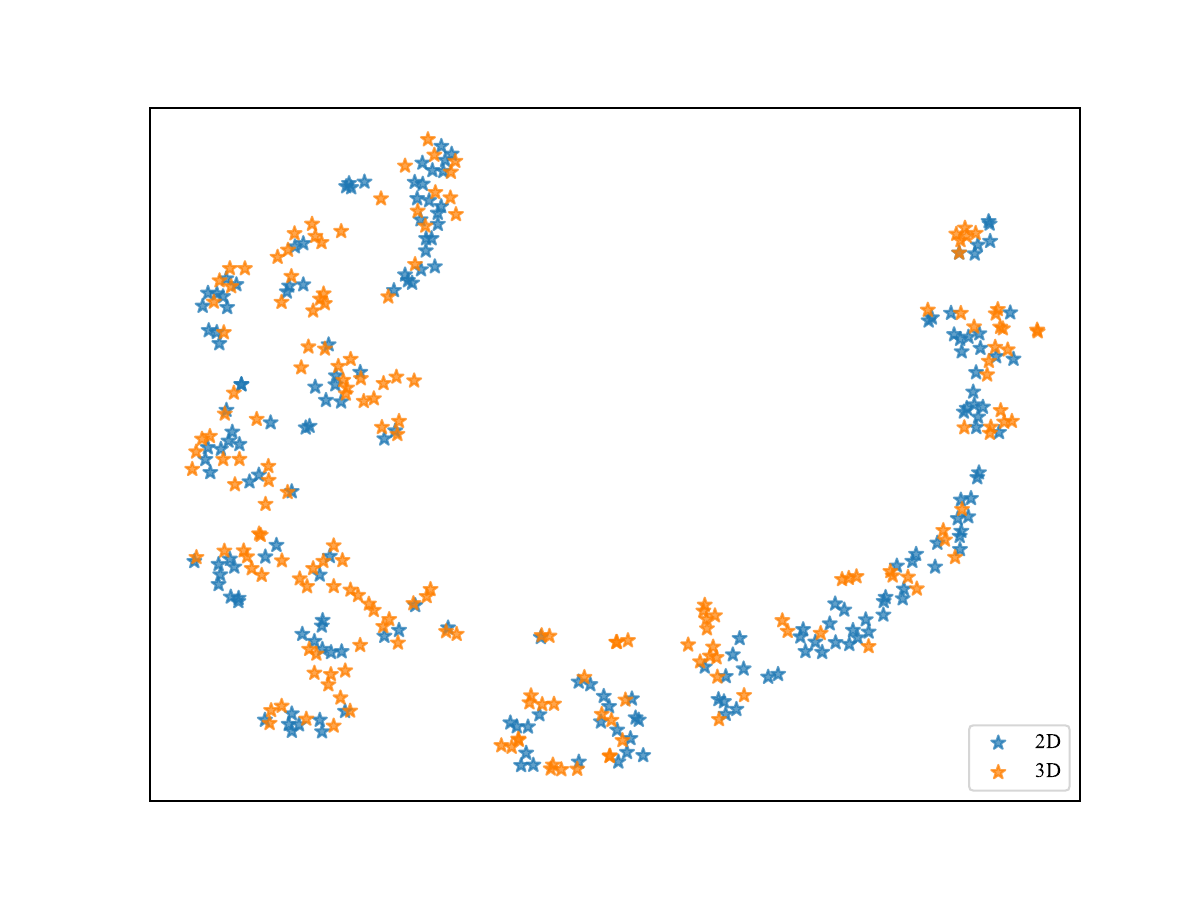} 
        \caption{t-SNE visualization of report representations by a pre-trained  BioClinicalBERT~\cite{alsentzer2019publicly}. Implementation details are in \textbf{Appendix A}.}  
        \label{fig:tsne_report}
    \end{subfigure} 
\caption{Observations that motivate the language-guided strategy for integrating 2D and 3D medical images in VLP.}
\label{fig:motivation}
\end{figure}

On the above aspect, the inherent heterogeneity of medical imaging from different modalities obstructs their effective collaboration and integration. One obvious and important problem is that medical images have different dimensions. For example, X-rays are 2D images, while CT scans are 3D images.
Nevertheless, when looking at their reports, which provide supervisions for the visual representation learning in VLP, it can be noticed that medical reports of images with different dimensions generally reflects same contents, which are an individual's health status.
As shown in Fig.~\ref{fig:example}, two reports mainly list the same abnormalities, and the X-ray and CT scan can contribute to a comprehensive understanding of pneumonia.
The rationale behind is that \textit{despite big differences, medical images from various modalities are all used for capturing the underlying features of an individual's health status, and such status are reflected in medical reports via language}.
Furthermore, we visualize the representation space of reports of 2D and 3D images in Fig.~\ref{fig:tsne_report}, showing that their representations lie together in a common semantic space.
This observation motivate us to \textit{map data from various medical image modalities into the shared semantic space, which is guided by language in reports}. This strategy not only tackles data-related issues but also fosters synergy and collaboration among distinct modalities, ultimately resulting in a more holistic understanding of an individual's health condition.

\begin{figure} [t] 
    \centering  
    \begin{subfigure}[b]{0.32\textwidth}   
        \centering  
        \includegraphics[width=\textwidth]{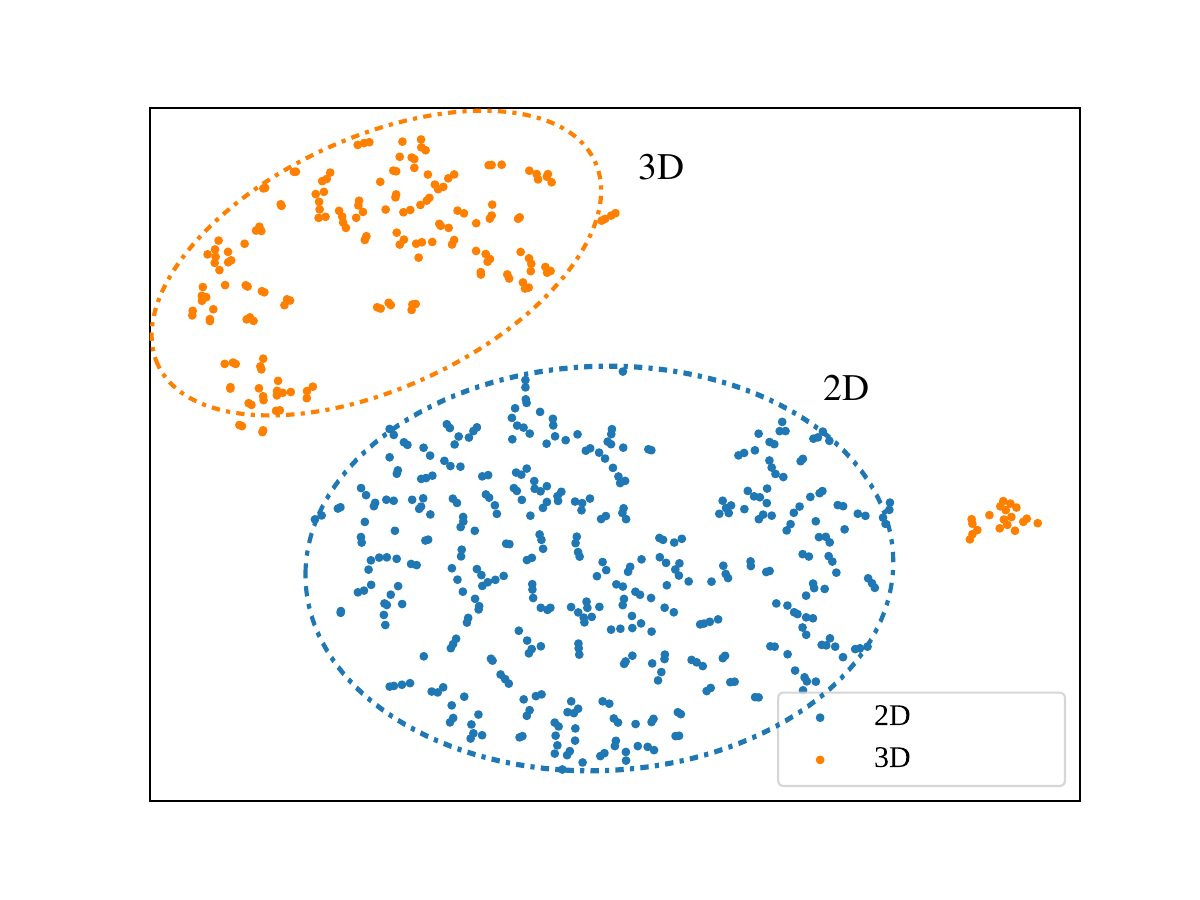}   
        \caption{}  
        \label{fig:tsnea}
    \end{subfigure}  
    \begin{subfigure}[b]{0.32\textwidth}   
        \centering  
        \includegraphics[width=\textwidth]{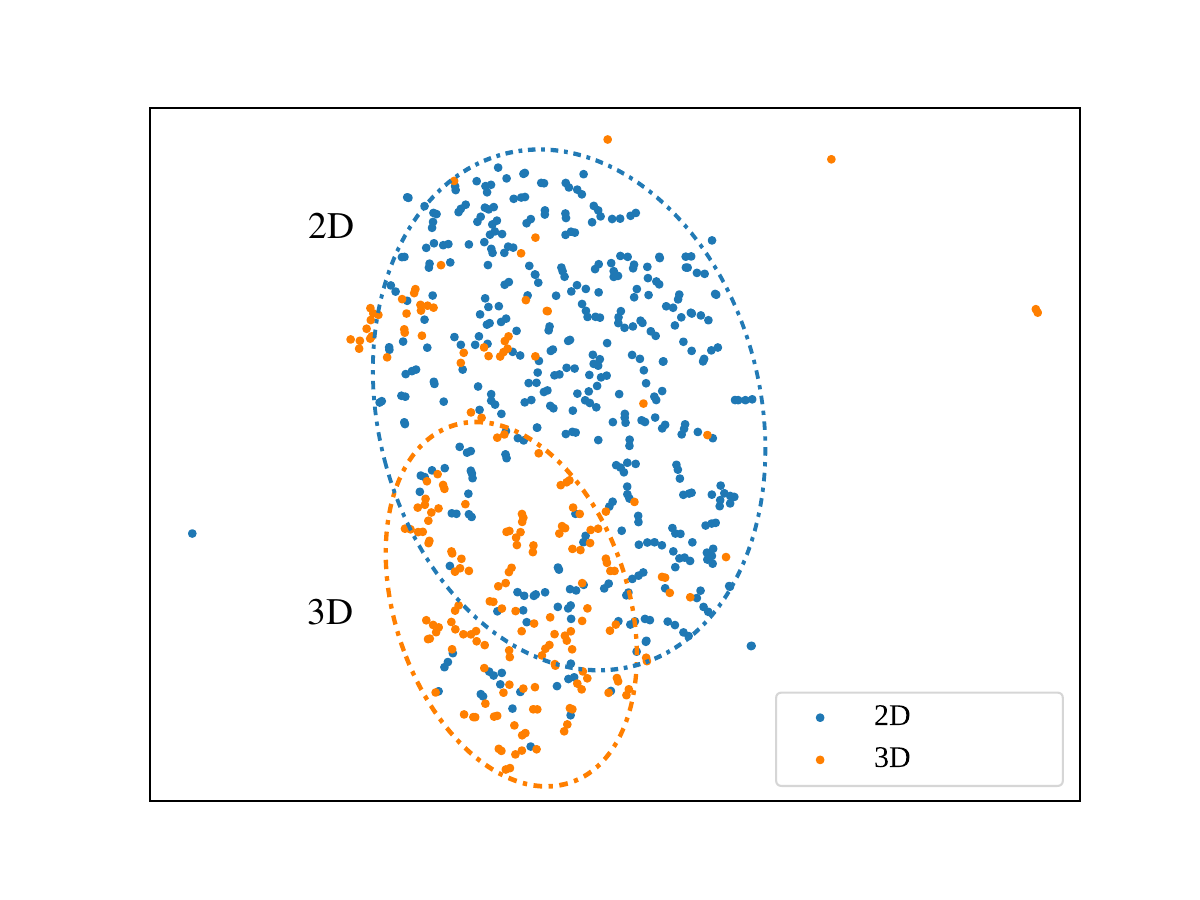} 
        \caption{}  
        \label{fig:tsneb}
    \end{subfigure}  
    \begin{subfigure}[b]{0.32\textwidth}   
        \centering  
        \includegraphics[width=\textwidth]{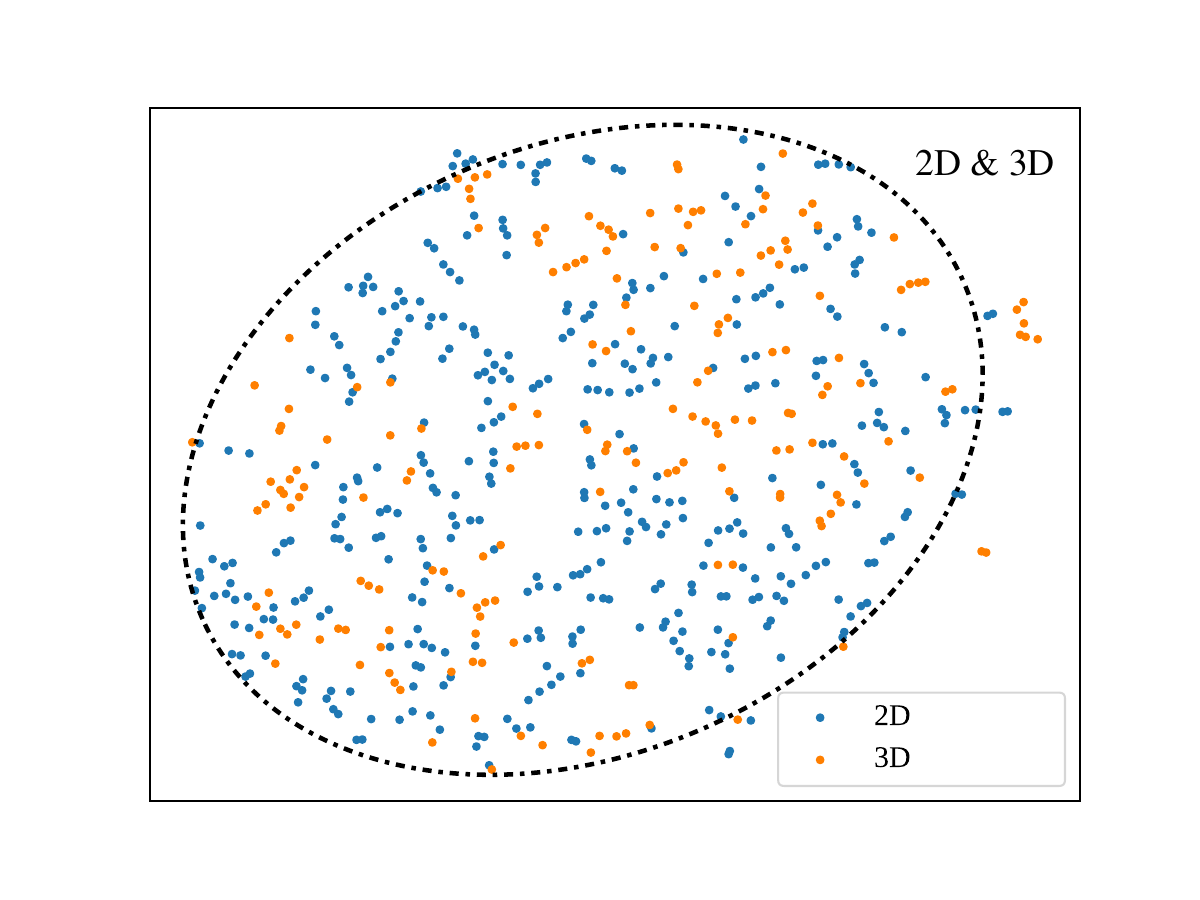}    
        \caption{}  
        \label{fig:tsnec}
    \end{subfigure}  
    \caption{\textcolor{black}{t-SNE visualizations of image representations by models trained with different VLP methods (2D: X-rays, 3D: CT). Both modalities are on the same disease, i.e., pneumonia, and the differences among sub-figures are highlighted with circles.  (a) Two models for different image modalities are trained individually in separate VLP process. (b) One models for different image modalities are trained in one VLP processes, but without designs in \ourmethod{}. (c) \ourmethod{}, introducing pseudo-paired `2D' and 3D images in one unified framework.
    }}
\label{fig:tsne}
\end{figure}

However, under the dilemma that paired 2D and 3D medical images with reports are unavailable, creating a unified model that effectively maps data of different modalities into a common space for combined learning is challenging, even with language guidance in reports. Fig.~\ref{fig:tsnea} demonstrates the representation space of two distinct modalities with different dimensions (i.e., 2D X-rays and 3D CT scans) when trained individually via VLPs. They are far apart in the representation space, even with same pathological information in reports. Furthermore, Fig.~\ref{fig:tsneb} shows simply unifying them in one model does not solve the problem. Although the distance between representations of two modalities are shortened to some extent, the degree of fusion is still far lower than that of report representations, which provides supervisions in VLP.


To address the above challenge, we propose \ourmethod{}, a novel \textbf{Uni}fied VL framework, designed to effectively integrate \textbf{Med}ical multi-modal \textbf{I}mages into a language-guided common semantic space, making them better help with each other. 
Notably, \ourmethod{} introduces a technique for creating the pseudo-pairs, enabling inputting both `2D' and 3D images in one forward pass. The pseudo-pair is created via an attentive selection method that can accurately identify text-relevant 2D slices without extra annotations.
This builds a data bridge between 2D and 3D medical images.
Besides, our unified VL framework contains one model and one VLP for different dimensional image data, bringing both 3D data and selected 2D slices closer to the same report representation space.
Moreover, a self-distillation technique is employed for the auxiliary masking and recovering task, further enhancing the interactions between 2D and 3D data within the image space.
Fig.~\ref{fig:tsnec} shows \ourmethod{} significantly reduces the distance between 2D and 3D features after undergoing our effective designs for cross-dimension pre-training, making it more like the report representation space.
Our experiment results (see Fig.~\ref{fig5}) also verify that \ourmethod{} can make 2D and 3D images help with each other, making most of them in the pre-training stage.

We conduct extensive experiments on several real-world medical datasets and various downstream tasks (i.e., classification, segmentation and retrieval).
The consistently superior performance well demonstrates the effectiveness and capacity of our \ourmethod{} framework, regardless of whether it is applied to full-scale data or limited data scenarios.
We also provide visualizations on slices selected by \ourmethod{}, verifying our claim that \ourmethod{} can identify key information from 3D medical images and create pseudo-pairs for integration.


\section{Related Work}

\subsubsection{Medical Self-Supervised Learning}
In the domain of medical image analysis, a number of self-Supervised learning (SSL) techniques have been developed to exploit the unique characteristics of medical data. These methods construct feature embedding spaces by designing pre-text tasks, such as solving jigsaw puzzles~\cite{noroozi1603unsupervised} and inpainting tasks~\cite{pathak2016context}. Recently, researchers have explored the use of 3D convolutional neural network (CNN) architectures while retaining established SSL tasks on 2D CNNs~\cite{tang2022self}. However, the diversity of medical data poses a significant challenge, as the development of a unified visual representation that adequately captures the intricacies of different data types remains a crucial yet complex task that requires further investigation. To address this challenge, \cite{xie2022unimiss} proposed UniMiSS, a universal medical self-supervised representation learning framework that overcomes the dimensionality barrier. Furthermore, \cite{nguyen2023joint} introduced Joint, an SSL framework capable of accommodating various data dimensions and generating versatile pre-trained weights for both 2D and 3D downstream applications. These approaches have made notable contributions to handling data from different modalities. However, they have given relatively less attention to the relationships and connections between different types of medical data.

\subsubsection{Medical Vision-Language Pre-Training}
Recently, Medical Vision-Language Pre-Training (VLP) has emerged as a promising approach for learning medical visual representations by leveraging naturally occurring paired descriptive text~\cite{zhang2022contrastive}. \cite{huang2021gloria} propose Gloria, an attention-based framework that contrasts image sub-regions and words in the paired report to learn global and local representations. \cite{wang2022multi} further optimize the framework from the perspective of disease in their method MGCA. These methods exhibit remarkable performance in various downstream tasks involving medical images. However, the application of medical VLP is primarily limited to 2D images, mainly due to the limited availability of extensive 3D medical image-text datasets. Compared to 2D medical image-text pairs, 3D images and reports contain more abundant information, which offers clear advantages for learning visual representations. While some methods~\cite{liu2023clip, chen2023generative} attempt to address this limitation by converting 3D data into 2D slices and subsequently employing generative models to generate captions for 3D medical data, this approach results in a loss of the original 3D volume structure information. Therefore, it is imperative to develop strategies that can effectively harness the valuable information present in 3D images and reports while preserving the structural integrity of the data. This will facilitate the enhancement of the learning process for visual representations in medical VLP.

\section{Methodology}

To effectively integrate multi-modal medical images into a language-guided common semantic space, we propose \ourmethod{} framework and present its components in this section.
We first show the overall pipeline and outline the key designs in Section~\ref{sec:met_pipeline}, and then explain details of each design in the following sub-sections. 

\subsection{Pipeline and Key Designs}
\label{sec:met_pipeline}

\begin{figure}[t]
    \includegraphics[width=\textwidth]{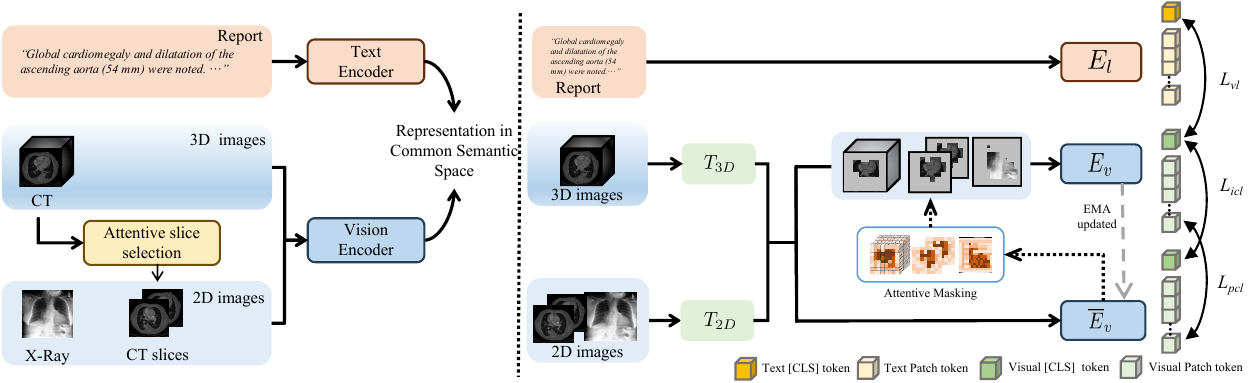} 
    \caption{\textcolor{black}{Illustration for the proposed \ourmethod{} framework. The overall pipeline is shown in the left part, and key designs are displayed in the right part. 
    }}
\label{fig:arch}
\end{figure}

The illustration of \ourmethod{} is in Fig.~\ref{fig:arch}.
Generally, \ourmethod{} is a vision-language pre-training framework, in which medical images and their reports are encoded by two encoders, separately, and they are jointly learned via the VL contrastive learning. The distinction of \ourmethod{} is that it can effectively taking both 2D and 3D images in an unified way, tackling data scarcity issue in the medical domain and making them help with each other.

\textit{First}, to overcome the challenges that no paired 2D and 3D image data exist, \ourmethod{} introduces a novel attentive slice selection strategy for creating pseudo-pairs. Here, when the input is a 3D volume, a portion of 2D slices which most relevant to the report are extracted from it, and then the selected slices are regarded as 2D image, thus forming pseudo 2D-3D pairs. Those selected 2D slices are fed into the network together with the original 3D volume in a forward pass, allowing jointly learning the relationships between 2D, 3D, and radiology reports, and ultimately form a unified feature space.
When the input is a 2D image, the slice selection process is omitted.
\textit{After that}, one vision encoder maps all multi-modal images, including original 2D and 3D ones and also selected 2D slices into the representation space.
This vision encoder has tokenizers $T_{2D}$ and $T_{3D}$ for 2D and 3D images, respectively, and a shared backbone $E_v$ for enabling better integration.
\textit{Then}, the model which contains the vision encoder and text encoder $E_l$ are learned end-to-end in one VLP via contrastive learning loss $L_{vl}$. Thus, 2D and 3D images can all be encoded into a common semantic space supervised by language information in reports.
\textit{Finally}, \ourmethod{} further introduces an auxiliary masking and recovering task for enhancing interactions among 2D and 3D tokens. This is completed via the self-distillation technique.

Our proposed attentive slice selection method is demonstrated in Section~\ref{section4.1}.
The VL contrastive learning used in our framework is explained in Section~\ref{section4.2}. 
The self-distillation technique is elaborated in Section~\ref{section4.3}.


\subsection{Creating Pseudo-Pairs via Attentive Slice Selection}
\label{section4.1}

According to our observation, we choose language in reports as the bridge to construct a cross-modal unified representation space.
Therefore, our goal is to select 2D slices that relevant most with reports from 3D volume. Note that this process is similar to how doctors view CT scans. They also base their report descriptions on some important slices.

\begin{figure}[t]
\centering
\includegraphics[width=1 \textwidth]{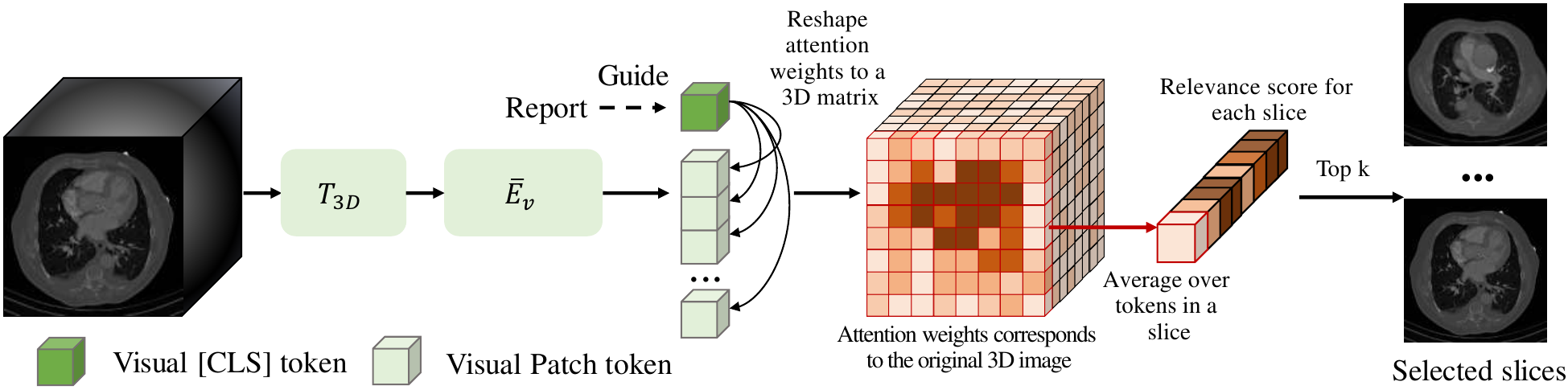}
\caption{Illustration for the attentive slice selection strategy that creates pseudo-pairs for 2D and 3D medical images.
}
\label{fig:slices}
\end{figure}

To achieve the goal, we design a novel attentive slice selection strategy which is guided by language, and show it in Fig.~\ref{fig:slices}. Basically, the relevance scores to the report are calculated for each slice and top $k$ slices with highest scores are selected.
Here, the attention value between the $\left[CLS\right]$ token produced by the vision encoder and other visual tokens are used as the basis for calculating relevance scores. As common VL pre-training methods \cite{radford2021learning, dong2023maskclip, ACLIP, li2023scaling}, this $\left[CLS\right]$ token is directly supervised by language, reflecting the most important lesion information in reports. Therefore, this token is likely to have large attention values with lesion-related 2D slices, and thus can be used for selection. Formally, the attention value vector $\mathbf{v} \in \mathbb{R}^{P}$ between $\left[CLS\right]$ token and $P$ image patch tokens is denoted as:
\begin{equation}\label{eq:Attn}
\mathbf{v}
=\frac{1}{HL} \sum_{l=1}^{L} \sum_{h=1}^{H} \operatorname{Softmax}\left(\frac{\mathbf{q}_{lh}({\left[CLS\right]}){{\mathbf{K}_{lh}(\left[Patch\right])}}}{\sqrt{C}}\right).
\end{equation}
There are $L$ layers with $l$ as the layer index and $H$ heads with $h$ as the head index. $\mathbf{q}_{lh}(\left[CLS\right]) \in \mathbb{R}^C$ denotes the query embedding of the $\left [ CLS \right ]$ token at layer $l$ and head $h$. $\mathbf{K}_{lh}(\left[Patch\right]) \in \mathbb{R}^{P \times C}$ denotes the key embedding of the all image patch tokens at layer $l$ and head $h$. $C$ is the dimension of the query and key embedding.



After obtaining the attention value for each token, which corresponds to a block of voxels in a slice of the original 3D image, the score for each slice is calculated by averaging values of all tokens in the slice. Therefore, the relevance score $s_i$ for the $i$-th slice is 
\begin{equation}\label{eq:slices_attn}
s_i
=\frac{1}{N} \sum_{j=1}^{N} \mathbf{v}_{ij}.
\end{equation}
Here, $\mathbf{v}_{ij}$ is the attention value for the $j$-th token in $i$-th slice that calculated via Eq~\ref{eq:Attn}. $N$ represents the number of tokens in a slice.
Hence, the relevance score of each slice to the report is obtained via the above average aggregating, and top $k$ slices are selected to establish a connection between 3D and 2D images, allowing learning a shared semantic feature space.

\subsection{Enabling Language Supervisions via VL Contrastive Learning}
\label{section4.2}

Generally, \ourmethod{} employs the VL contrastive learning loss in CLIP~\cite{CLIP} to introduce language supervisions in reports for learning representations of multi-modal medical images, including both 2D and 3D ones, and build the alignment between language and images.
For 2D images (i.e., X-rays), we directly use $T_{2D}$ and $E_v$ for feature extraction, obtaining the $\left[CLS\right]$ token that contains global image information, and then align it with the $\left[CLS\right]$ token generated by the language encoder $E_l$ that contains global report information.
For 3D images (i.e., CT scans), we first select 2D slices related to the report from the 3D volume via the strategy described in Section~\ref{section4.1}. Then, 2D slices and the original 3D volume are tokenized by $T_{2D}$ and $T_{3D}$, respectively. All tokens are concatenated and processed together by $E_v$. Finally, same as 2D images, the contrastive learning loss $L_{vl}$ is applied to image $\left[CLS\right]$ tokens and language $\left[CLS\right]$ tokens.

A highlight of \ourmethod{} is the synergistic effect of the attentive slice selection strategy and the VL contrastive learning. 
\textit{On the one hand}, VL contrastive learning enables language supervisions, which is directly applied to the visual $\left[CLS\right]$ token, and thus this token contains important information in reports. This is the reason why attention weights of the visual $\left[CLS\right]$ token are utilized as the basis for 2D slice selection. Only in this way can chosen 2D slices carry the supervision information from the report and, together with the 3D features, construct a joint feature space. Here, the key insight is language is used as the coordinate for learning a common semantic space.
\textit{On the other hand}, the attentive slice selection make 2D and 3D feature space more integrated, even without paired data. Thus, they can benefit more from each other, and this common space can amplify similar information between medical images and reports, as they are all about an individual's health status. In this way, the alignment between images and reports is impelled.
\textit{In summary}, this two designs bring together representations of both 2D and 3D medical images, and make them close to the report representation space at the same time, achieving the effect of one plus one greater than two on building the common semantic space.



\subsection{Enhancing Cross-Dimension Interactions via Self-distillation}
\label{section4.3}

As medical images themselves are multi-modal and also share some common information, to make the most of these information, we introduce an auxiliary task in which tokens of 2D and 3D images can communicate with each other. Therefore, cross-dimension interactions are reinforced in this way, further enhancing the integration of 2D and 3D images.

We adopt a simple and straightforward auxiliary task design, which is masking and recovering, and use the self-distillation method to complete the task~\cite{ACLIP, zhou2021ibot}, due to its simplicity and effectiveness. Specifically, during the training process, we maintain an online student network $E_v$, and a teacher network $\overline{E}_v$, which is updated by exponential moving averaged (EMA) over $E_v$. The teacher network provides the distillation target as follows. We mask a large proportion of both 2D and 3D image tokens for inputs to the student network, while keeping them complete for inputs to the teacher network. Then, the student network is required to output similar features as the teacher network, even if a significant amount of information is missed in its inputs. 

The head and loss of DINO~\cite{caron2021emerging} is used for implementing self-distillation here. This loss is applied to both the global $\left[CLS\right]$ tokens (i.e., image contrastive learning loss $L_{icl}$) and all local patch tokens (i.e., patch contrastive learning loss $L_{pcl}$), thereby enabling interactions at different levels of granularity to enhance feature robustness.
Besides, since language guidance is the key to realize the common semantic space, the mask strategy here is also based on attention weights, similar to the slice selection method. Tokens with highest attention weights are retained.

Since a large ratio of inputs are masked, this non-trivial auxiliary task forces the model to dig more cross-dimension information, implicitly requiring the model to recover 2D images from 3D ones and vice versa. Therefore, 2D and 3D image tokens can communicate with each other, and cross-dimension interactions are enhanced. Besides, the language-guided masking strategy also makes the interactions more meaningful.

\section{Experiments}


\subsection{Data and Settings in the Pre-Training Stage}
\label{4.1}
\subsubsection{Dataset} \quad We pre-train our \ourmethod{} framework on the JPG version of 2D X-rays dataset MIMIC-CXR 2.0.0 \cite{johnson2019mimic} and 3D CT scans dataset BIMCV \cite{mincv}. Specially, for the preprocessing of 2D datasets, we eliminate all side-view images to align with downstream tasks that solely employ front-view images. Additionally, we discard any brief reports that contain fewer than three sentences for 2D and 3D dataset to preserve the integrity of our dataset.
 
\subsubsection{Implementation Details} \quad We utilize ViT-B/16 \cite{dosovitskiy2020image} as the vision encoder to extract representations in the common feature space for 2D and 3D visual data. For text feature encoding, we use BioClinicalBERT \cite{alsentzer2019publicly} as the text encoder to obtain the report embeddings. The vision encoder and text encoder are universal among 2D X-rays and 3D CT scans data. It is worth noting that the patch embed module of vision encoder has different operations for 2D X-rays and 3D CT scans. In general, the image size of 2D images is $224 \times 224$ and the volume size of 3D volumes is $128 \times 128 \times 32$. We pre-train our \ourmethod{} framework 50 epochs on 8 Tesla V100 GPUs with batch size of 144. The optimizer is AdamW \cite{loshchilov2017decoupled} with learning rate of $2e^{-5}$ and weight decay of 0.05, where the learning rate follows a linear warmup with cosine annealing scheduler \cite{loshchilov2016sgdr}. We initialize learning rate as $1e^{-8}$ and warmup epoch as 20.



\subsection{Downstream Tasks for Evaluation}
\label{4.2}

\subsubsection{Medical Image Classification} \quad  We conduct medical image classification on both 2D and 3D datasets. First, we explore three representative 2D datasets: (1) CheXpert \cite{irvin2019chexpert}, which contains 191,229 frontal-view chest radiographs. The task is to classify each image into 5 individual binary labels: atelectasis, cardiomegaly, consolidation, edema, and pleural effusion. Following \cite{zhang2022contrastive, huang2021gloria}, we hold out the expert-labeled validation set as test data and randomly select 5,000 radiographs from training data for validation. (2) RSNA Pneumonia \cite{shih2019augmenting}. \textcolor{black}{We use the stage 2 version, which contains around 29,700 frontal view chest radiographs.} The task is a binary classification, \emph{i.e.}, classifying each chest image into normal or pneumothorax positive. Following \cite{huang2021gloria}, we manually split the dataset into training, validation, and test set with 70\%/15\%/15\% ratio. (3) COVIDx \cite{wang2020covid}, which contains over 30\emph{k} CXR images from a multinational cohort of over 16,600 patients. This dataset contains 16,490 positive COVID-19 images from over 2,800 patients. We use the latest version 6 of this dataset. The task is a three-class classification, \emph{i.e.}, classifying each radiograph into COVID-19, non-COVID pneumonia or normal. We use the original validation dataset as test data and manually split 10\% of original training set for validation.

We conduct medical volume classification on two representative datasets: CC-CCII \cite{zhang2020clinically} and LUNA 16 \cite{setio2017validation}. (1) CC-CCII, we use a clean version named Clean-CC-CCII, which consists of 340,190 slices of 3,993 scans from 2,698 patients. The task is to classify each volume into three categories: novel coronavirus pneumonia (NCP), common pneumonia (CP), and normal (Normal). (2) LUNA 16 \cite{setio2017validation}, which is established from LIDC-IDRI \cite{armato2011lung}. The dataset contains 888 CT scans with annotations which removes CT scans with slice thickness greater than 3mm of LIDC-IDRI database. The task is a binary classification,  \emph{i.e.}, classifying each CT volume into pulmonary nodule or normal. 

Linear classification is employed to assess the representational capacity of our universal vision-language pre-training model. This model involves keeping the pre-trained ViT vision encoder fixed and solely training a linear classification head initialized randomly for the classification task with varying amounts of training data on each dataset. In addition to this, we perform full model fine-tuning to assess the effectiveness of \ourmethod{} on 3D CT scans.

\subsubsection{Medical Image Semantic Segmentation} \quad To assess the effectiveness of \ourmethod{} for medical semantic segmentation, we carry out experiments across a range of representative 2D and 3D datasets. First, we conduct experiments on a representative 2D dataset. RSNA Pneumonia \cite{shih2019augmenting}, contains 29700 frontal view radiograph. The task is to predict bounding boxes indicating evidence of pneumonia. We randomly split the original training set into 16,010/5,337/5,337 for training/validation/testing. We convert object detection ground truths into masks for semantic segmentation. (2) BCV \cite{landman2015miccai}, which consists of 50 CT scans. We partition the training and testing sets following \cite{xie2022unimiss}. 

We evaluate the segmentation performance
with the paradigm that we use the pre-trained vision encoder as a frozen encoder and train a decoder portion using 1\%, 10\% and 100\% training data on RSNA dataset and 20\%, 40\%, 100\% training data on BCV dataset. Dice scores are reported to evaluate the segmentation performance. 

\subsection{Compared Methods}

To evaluate the effectiveness of \ourmethod{}, we conduct comparative analyses with various medical VLP methods (ConVIRT \cite{zhang2022contrastive}, GLoRIA \cite{huang2021gloria}, MGCA \cite{wang2022multi}, LOVT \cite{muller2022joint}, PRIOR \cite{cheng2023prior}) . However, it is important to note that these methods, different from ours, are tailored specifically for X-ray and their corresponding medical reports. In addition to this, we have also compared our method with several 2D and 3D co-learning methods \cite{xie2022unimiss, nguyen2023joint}. UniMiSS not only learns 2D and 3D representations but also concurrently learn all 2D sections derived from 3D volumes, along with all 2D X-ray data. Joint directly learn all 2D sections derived from 3D volumes, along with all 2D X-ray data through contrastive learning.

\begin{table}[t]
    \centering
    \caption{2D Linear classification results on CheXpert, RSNA and COVIDx with 1\%, 10\%, 100\% training data. Area under ROC curve (AUROC [\%]) are reported for CheXpert and RSNA dataset, and accuracy (ACC [\%]) is reported for COVIDx dataset. The best results are highlighted in \textbf{boldface}.}
    \setlength{\tabcolsep}{0.8mm}
    \begin{tabular}{l c c c c c c c c c} 
        \hline
        {} & \multicolumn{3}{c}{CheXpert(AUC)} & \multicolumn{3}{c}{RSNA(AUC)} & \multicolumn{3}{c}{COVIDx(ACC)} \\
        Method & 1\% & 10\% & 100\% & 1\% & 10\% & 100\% & 1\% & 10\% & 100\% \\
        \hline
        Random Init & 56.1 & 62.6 & 65.7 & 58.9 & 69.4 & 74.1 & 50.5 & 60.3 & 70.0 \\
        ImageNet Init & 74.4 & 79.9 & 81.4 & 74.9 & 74.5 & 76.3 & 64.8 & 78.8 & 86.3 \\
        \hline
        \hline
        \emph{pre-trained on CheXpert} \\
        DSVE\textcolor{black}{~\cite{engilberge2018finding}} & 50.1 & 51.0 & 51.5 & 49.7 & 52.1 & 57.8 & - & - & - \\
        VSE++\textcolor{black}{~\cite{faghri2017vse++}} & 50.3 & 51.2 & 52.4 & 49.4 & 57.2 & 57.9 & - & - & - \\
        GLoRIA\textcolor{black}{~\cite{huang2021gloria}} & 86.6 & 87.8 & 88.1 & 86.1 & 88.0 & 88.6 & 67.3 & 77.8 & 89.0 \\
        \hline
        \hline
        \emph{pre-trained on MIMIC-CXR} \\
        Caption-Transformer\textcolor{black}{~\cite{cornia2020meshed}} & 77.2 & 82.6 & 83.9 & - & - & - & - & - & - \\
        Caption-LSTM\textcolor{black}{~\cite{xu2015show}} & 85.2 & 85.3 & 86.2 & - & - & - & - & - & - \\
        Contrastive-Binary\textcolor{black}{~\cite{tan2019lxmert}} & 84.5 & 85.6 & 85.8 & - & - & - & - & - & - \\
        ConVIRT\textcolor{black}{~\cite{zhang2022contrastive}} & 85.9 & 86.8 & 87.3 & 77.4 & 80.1 & 81.3 & 72.5 & 82.5 & 92.0 \\
        GLoRIA-MIMIC\textcolor{black}{~\cite{huang2021gloria}} & 87.1 & 88.7 & 88.0 & 87.0 & 89.4 & 90.2 & 66.5 & 80.5 & 88.8 \\
        LOVT~\cite{muller2022joint} & 85.1 & 88.1 & 88.3 & 85.5 & 86.5 & 89.3 & - & - &- \\
        PIROR~\cite{cheng2023prior} & 86.2 & 88.3 & 88.6 & 85.7 & 87.1 & 89.2 & - & - & - \\
        MGCA (ResNet-50)\textcolor{black}{~\cite{wang2022multi}} & 87.6 & 88.0 & 88.2 & 88.6 & 89.1 & 89.9 & 72.0 & 83.5 & 90.5 \\
        MGCA (ViT-B/16)\textcolor{black}{~\cite{wang2022multi}} & \textcolor{black}{88.8} & \textcolor{black}{89.1} & \textcolor{black}{89.7} & \textcolor{black}{89.1} & \textcolor{black}{89.9} & \textcolor{black}{90.8} & \textcolor{black}{74.8} & \textcolor{black}{84.8} & \textcolor{black}{92.3} \\
        \hline
        \hline
         \rowcolor{graycolor} {\textbf{\ourmethod{} (ViT-B/16)}} & {\textbf{89.4}} & {\textbf{89.7}} & {\textbf{90.5}} & {\textbf{90.0}} & {\textbf{90.4}} & {\textbf{91.5}} & {\textbf{80.3}} & {\textbf{92.4}} & {\textbf{94.6}} \\
        \hline
    \end{tabular}
    \label{table1}
\end{table}

\begin{table}[t]  
  \begin{minipage}{0.46\linewidth}  
    \centering
    \caption{3D Linear classification results on CC-CCII with 1\%, 10\%, 100\% training data. Accuracy are reported for the dataset. $\ast$ denotes the input size $16\times96\times96$. Others is $32\times128\times128$. The best results are highlighted in \textbf{boldface}.}
    \setlength{\tabcolsep}{1.6mm}
    \begin{tabular}{l c c c c c} 
        \hline
        {} & \multicolumn{3}{c}{CC-CCII} \\
        Method & 1\% & 10\% & 100\% \\
        \hline
        Random Init & {43.4} & {69.7} & {74.8} \\
        \hline
        \hline
        {UniMiSS$^\ast$}\textcolor{black}{~\cite{xie2022unimiss}} & {41.6} & {73.1} & {84.1} \\
        \hline
        \hline
        \rowcolor{graycolor} {\ourmethod{}$^\ast$} & {64.2} & {75.1} & {84.9} \\ 
        \rowcolor{graycolor} {\textbf{\ourmethod{}}} & \textbf{75.6} & \textbf{84.8} & \textbf{89.4} \\
        \hline
    \end{tabular}
    \label{table2}
  \end{minipage}
  \hspace{0.2cm}
  \begin{minipage}{0.47\linewidth} 
    \setlength{\tabcolsep}{0.8mm}
    \centering
    \caption{3D Fine-tuning results on CC-CCII, RICORD with full training data. ACC [\%] is reported for CC-CCII and AUC [\%] is reported for LUNA2016-v2. The best results are highlighted in \textbf{boldface}.}
    \begin{tabular}{l c c} 
        \hline
        {Method} & CC-CCII & LUNA \\
        \hline 
        \emph{supervised} \\
        {ResNet3D101} & {85.5} & {-}  \\
        {CovidNet3D-L} & {88.7} & {-} \\
        \hline
        \hline
        \emph{unsupervised} \\
        {Joint}\textcolor{black}{~\cite{nguyen2023joint}} & {-}  & {94.2} \\
        \hline
        \hline
        \rowcolor{graycolor} \textbf{\ourmethod{}} & \textbf{93.8} & \textbf{95.9} \\
        \hline
    \end{tabular}
    \centering
    \label{table3}
  \end{minipage}
\end{table}

\subsection{Experiment Results}
\label{4.3}
\subsubsection{Results on 2D Medical Image Classification} \quad Table~\ref{table1} reports the results of linear classification on three 2D medical image classification datasets (CheXpert, RSNA and COVIDx). The results of other methods on CheXpert and RSNA are from original paper. The methods including \ourmethod{} shown in the table are pre-trained on MIMIC-CXR dataset, which achieves a fair comparison.  As for the state-of-the-art method, MGCA, we mainly compare the performance with the MGCA (ViT-B/16) which employs the ViT as the visual encoder. It is obvious that our method shows the best performance in the three 2D medical image classification for the different training data ratio (1\%, 10\%, 100\%), outperforming the state-of-the-art MGCA (ViT-B/16) by a large margin. Specifically, our method outperforms MGCA with ViT-B/16 backbone with +0.6\%, +0.6\%, +0.8\% AUROC on CheXpert dataset, +0.9\%, +0.5\%, +0.7\% AUROC on RSNA dataset and +5.5\%, +7.6\%, +2.3\% ACC on COVIDx dataset under the 1\%, 10\%, 100\% training ratio respectively. The significant improvement indicates the data efficiency and effectiveness of our method.

\subsubsection{Results on 3D Medical Volume Classification} \quad Table~\ref{table2} reports the results of linear classification on the 3D medical image classification dataset, CC-CCII. We compare \ourmethod{} with UniMiss~\cite{xie2022unimiss}. \textcolor{black}{To our knowledge, the UniMiSS \cite{xie2022unimiss} is the state-of-the-art unified method to process 2D and 3D medical images}. We show the performances of both UniMiSS and our \ourmethod{}, where the results are that our method achieves a +{22.6}\%, +{2.0}\% and +{0.8}\% ACC gain on CC-CCII dataset comparing with the UniMiSS under the 1\%, 10\%, 100\% training ratio respectively. The significant improvement indicates the data efficiency and effectiveness of our method.

\begin{table}[t]  
  \begin{minipage}{0.48\linewidth} 
    \centering
    \caption{2D Semantic segmentation results (Dice [\%]) on RSNA with 1\%, 10\% and 100\% training labels. }
    \setlength{\tabcolsep}{1pt}
    \begin{tabular}{l c c c} 
        \hline
        {} & \multicolumn{3}{c}{RSNA} \\
        {Method} & 1\% & 10\% & 100\% \\
        \midrule
        {ConVIRT} & {55.0} & {67.4} & {67.5} \\
        {GLoRIA} & {59.3} & {67.5} & {67.8} \\
        {GLoRIA-MIMIC} & {60.3} & {68.7} & {68.3} \\
        {MGCA} & {88.6} & {81.2} & {94.3} \\
        {MGCA (ViT-B/16)} & {66.2} & {71.3} & {73.6} \\
        \hline
        \rowcolor{graycolor} \textbf{\ourmethod{}} & \textbf{67.8} & \textbf{73.1} & \textbf{75.3} \\
        \hline
    \end{tabular}

    \label{table6}
  \end{minipage}
  \hspace{0.2cm}
  \begin{minipage}{0.48\linewidth}  
    \centering
    \caption{2D Object detection results (mAP [\%]) on RSNA with 1\%, 10\%, 100\% training data. }
    \setlength{\tabcolsep}{1pt}
    \begin{tabular}{l c c c } 
        \hline
        {} & \multicolumn{3}{c}{\textcolor{black}{RSNA}} \\
        \textcolor{black}{Method} & \textcolor{black}{1\%} & \textcolor{black}{10\%} & \textcolor{black}{100\%} \\
        \midrule
        \textcolor{black}{ConVIRT} & \textcolor{black}{8.2} & \textcolor{black}{5.6} & \textcolor{black}{17.9} \\
        \textcolor{black}{GLoRIA} & \textcolor{black}{9.8} & \textcolor{black}{14.8} &  \textcolor{black}{18.8} \\
        \textcolor{black}{GLoRIA-MIMIC} & \textcolor{black}{11.6} & \textcolor{black}{16.1} & \textcolor{black}{24.8} \\
        \textcolor{black}{MGCA} & \textcolor{black}{12.9} & \textcolor{black}{16.8} & \textcolor{black}{24.9} \\
        \textcolor{black}{MGCA (ViT-B/16)} & \textcolor{black}{14.7} & \textcolor{black}{18.4} & \textcolor{black}{25.8} \\
        \hline
        \rowcolor{graycolor} \textcolor{black}{\textbf{\ourmethod{}}} & \textcolor{black}{\bf 15.5} & \textcolor{black}{{\bf 19.2}} & \textcolor{black}{{\bf 26.6}} \\ 
        \hline
    \end{tabular}
    \label{table7}
  \end{minipage}
\end{table}

When fine-tuning the total vision encoder with full training data, as listed in Table~\ref{table3}, our method gets the best performance on the multiple 3D medical volume classification datasets (CC-CCII and LUNA) compared with other methods. It is observed that our method achieves with {93.8}\% ACC on CC-CCII dataset, and {95.9}\% ACC on LUNA2016-v2 dataset respectively. The remarkable performance of our method shows the generalization of our method for 2D and 3D medical classification tasks. It demonstrates our framework possesses the ability of extracting universal features for 3D CT scans. 

\subsubsection{Results on 2D Medical Semantic Segmentation}
We evaluate the impact of local representation by analyzing its performance in the task of semantic segmentation. Table~\ref{table6} presents the results of all methods on the RSNA dataset. \ourmethod{} significantly outperforms the current state-of-the-art algorithm, MGCA. When using 1\% training data, \ourmethod{} achieves 67.8\% Dice, surpasssing the MGCA 1.6\%. In addition to segmentation tasks, we also conducted detection tasks on the RSNA Pneumonia dataset to demonstrate the localized capabilities of \ourmethod{}. The objective is to identify and predict bounding boxes that signify the presence of pneumonia. Utilizing ViT-B as our backbone, we can create an effective feature pyramid from a single-scale feature map, which simplifies the architecture. For the detection task, we leveraged ViTDet \cite{li2022exploring} with a pre-trained ViT-B serving as a frozen backbone, while only the non-backbone layers were fine-tuned. To assess data efficiency, we fine-tuned the model with varying amounts of training data: 1\%, 10\%, and 100\%. Table~\ref{table7} presents the results of all methods on the RSNA dataset.

\subsubsection{Results on 3D Medical Semantic Segmentation} \quad Table~\ref{table8} showcases the performance of \ourmethod{} on the BCV dataset for semantic segmentation in 3D medical image. Compared to the baseline method UniMiSS, our approach yields a 0.6\% and 0.4\% increase in accuracy with limited label availability of 40\% and 100\%, respectively. These results underscore the superior capability of \ourmethod{} in extracting meaningful features and effectively capitalizing on limited annotated data. This performance gain reflects \ourmethod{}'s adeptness at harnessing local representations for the semantic segmentation task. 

\subsection{Visualization and Ablation Study}

\begin{figure}[t]
\centering
\includegraphics[width=1 \textwidth]{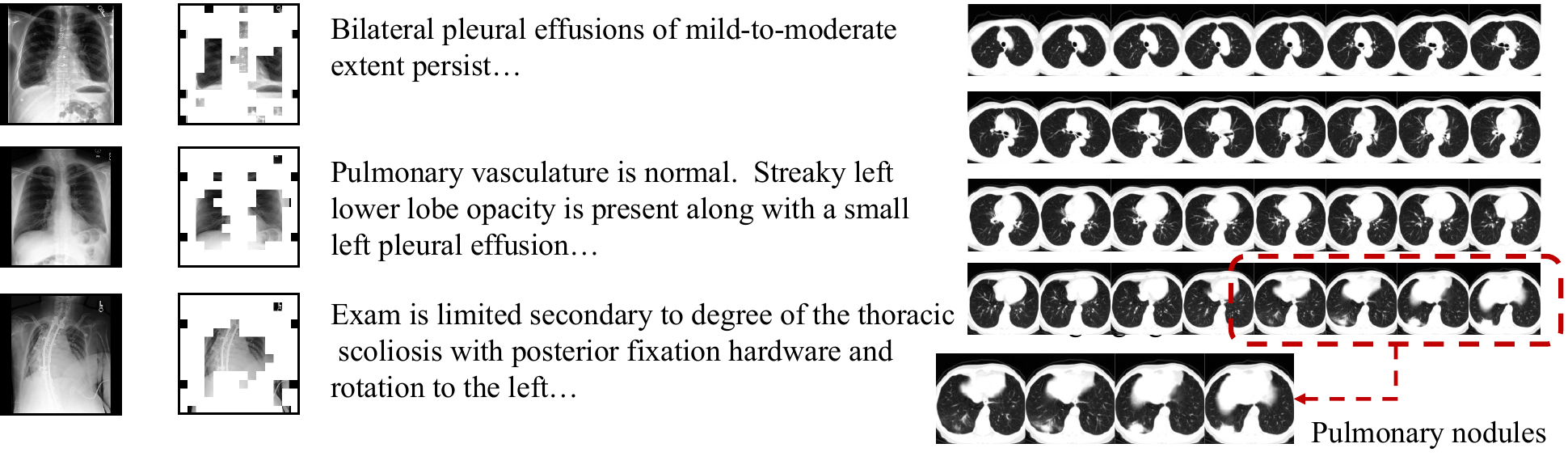}
\caption{Visualization of masking and slices selection result under the guidance of language.}
\label{fig6}
\end{figure}

\begin{wraptable}{r}{0.5\textwidth}
\vspace{-1.0cm}
\centering
    \caption{3D Semantic segmentation results (Dice [\%]) on BCV with 20\%, 40\% and 100\% training labels. Best results of each setting are in boldface.}
    \begin{tabular}{l c c c } 
        \hline
        {} & \multicolumn{3}{c}{BCV} \\
        Method & 20\% & 40\% & 100\% \\
        \hline
        {MoCo v3} & {74.5} & {78.2} & {82.0} \\
        {DINO} & {75.3} & {78.9} & {82.6} \\
        {UniMiSS} & \textbf{78.0} & {81.0} & {85.0} \\
        \hline
        \rowcolor{graycolor} \textbf{\ourmethod{}} & {77.5} & \textbf{81.6} & \textbf{85.4} \\
        \hline
    \end{tabular}
    \label{table8}
\end{wraptable}

\subsubsection{Visualization} To enhance the clarity and showcase the efficacy of our linguistically-informed selection process, we conduct a comprehensive visual presentation in Fig.~\ref{fig6}. This includes the display of the original X-ray images alongside the masked X-rays, where the regions of interest are prominently highlighted. Accompanying these images are the corresponding medical reports that provide context and diagnostic insights. Furthermore, we exhibit the original CT scans , as well as selected lesion slices. On the left side of Fig.~\ref{fig6}, the first row effectively demonstrates how \ourmethod{} accurately captures the areas referenced in the report, including the ``Normal post-operative alignment of the sternal wires'' and ``Bilateral pleural effusions of mild-to-moderate extent persist''. In addition, the second and third cases adeptly showcase the detection of pleural effusion and scoliosis, further emphasizing the method's precision. The right side of Fig.~\ref{fig6} displays the comprehensive slice selection process employed by \ourmethod{}. Amidst the extensive collection of CT scan slices, our method exhibits remarkable accuracy in pinpointing the slices containing lesions. As an example, the presence of pulmonary nodules is clearly noticeable in slices 28-31.

\begin{figure}[t]
\centering
    \begin{subfigure}[b]{0.24\textwidth}   
        \centering  
        \includegraphics[width=\textwidth]{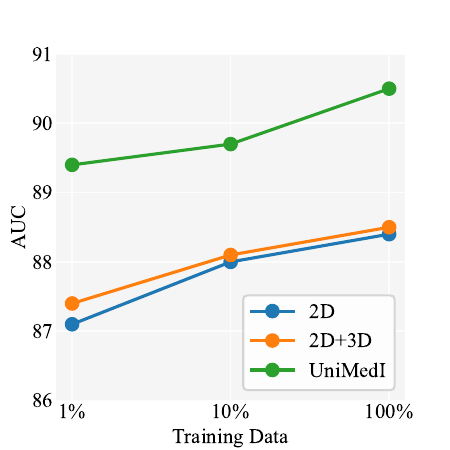}
        \caption{CheXpert Dataset.}  
    \end{subfigure}  
    \begin{subfigure}[b]{0.24\textwidth}  
        \centering  
        \includegraphics[width=\textwidth]{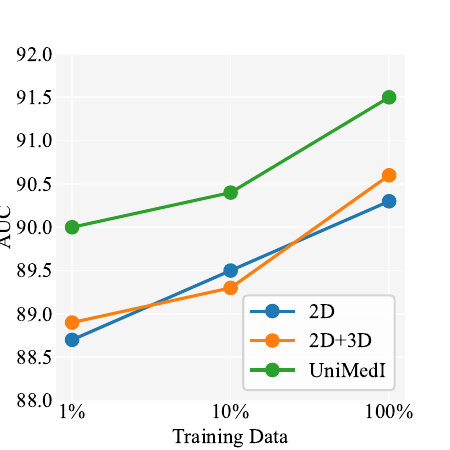} 
        \caption{RSNA Dataset. }  
    \end{subfigure} 
    \begin{subfigure}[b]{0.24\textwidth}  
        \centering  
        \includegraphics[width=\textwidth]{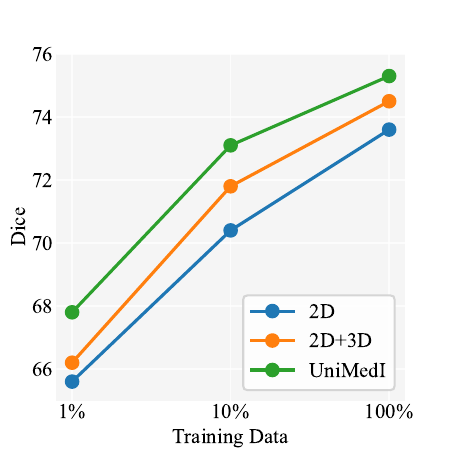} 
        \caption{RSNA Dataset. }  
    \end{subfigure} 
    \begin{subfigure}[b]{0.24\textwidth}  
        \centering  
        \includegraphics[width=\textwidth]{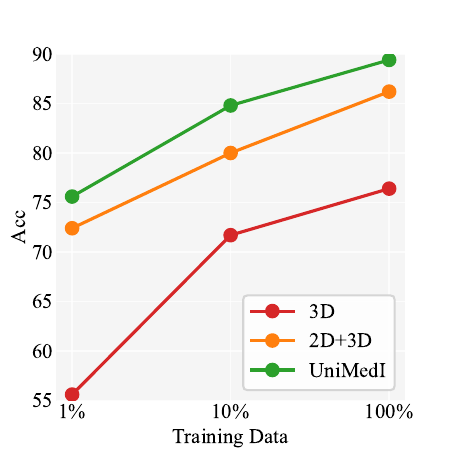} 
        \caption{CC-CCII Dataset. }  
    \end{subfigure} 
\caption{Comparative study of common semantic space versus separate. The results presented here are set in the context of linear classification (2D dataset CheXpert, RSNA and 3D dataset CC-CCII) settings. We report Area under ROC curve (AUROC [\%]) on (a) CheXpert Classification and (b) RSNA Classification, Dice [\%] on (c) RSNA Segmentation and Acc [\%] on (d) CC-CCII Classification.}.
\label{fig5}
\end{figure}

\subsubsection{Importance of the Common Semantic Space} 
As illustrated in Fig~\ref{fig:tsne}, we provide visualizations for three types of feature spaces to elaborate on the motivation behind \ourmethod{}. We now proceed to its quantitative analysis. As shown in Fig~\ref{fig5}, conducting joint training on both  modalities (2D $+$ 3D) within the same model shows certain improvements over single-modality training, but the enhancement is not significant.  The design of \ourmethod{}, through techniques like language-guided pseudo pairs and self distillation, enables better joint modeling of 2D and 3D information. This leads to substantial performance improvements across all settings, which cannot be achieved by simple adding 2D and 3D data together.

\setlength{\tabcolsep}{4pt}
\begin{table}[t]
    \caption{
    Ablation study of component design. We report Area under ROC curve (AUROC [\%]) on CheXpert and RSNA datasets, and (Acc [\%]) on CC-CCII dataset. $VL$ represents the default experiment setting include image-text contrastive loss $L_{vl}$ with random slices selection. $SD$ will include $L_{icl}$ and $L_{pcl}$ loss to execute Self-Distillation. $Attn$ will use attentive slices selection instead of random.
    Best results of each setting are in boldface.
    }
    \centering
    \begin{tabular}{c c c | c c c c c c | c c c c}
    \toprule
    \multicolumn{3}{c}{Training tasks} & \multicolumn{3}{c}{CheXpert (AUC)} & \multicolumn{3}{c}{RSNA (AUC)} & \multicolumn{3}{c}{CC-CCII (Acc)} \\
    $VL$ & $SD$ & $Attn$ & $1\%$ & $10\%$ & $100\%$ & $1\%$ & $10\%$ & $100\%$ & $1\%$ & $10\%$ & $100\%$ \\
    \midrule
    \checkmark & & & 87.4 & 88.1 & 88.5 & 88.9 & 89.3 & 90.6 & 72.4 & 80.0 & 86.2 \\
    \checkmark & \checkmark & & 89.0 & 89.3 & 90.1 & 89.5 & 90.1 & 91.2 & 74.6 & 80.9 & 86.7 \\
    \rowcolor{graycolor} \checkmark & \checkmark & \checkmark & \textbf{89.4} & \textbf{89.7} & \textbf{90.5} & \textbf{90.0} & \textbf{90.4} & \textbf{91.5} & \textbf{75.6} & \textbf{84.8} & \textbf{89.4} \\
    \bottomrule
    \end{tabular}
    \label{table5}
\end{table}

\subsubsection{Ablation on Self-Distillation and Attentive Slices Selection}
Results in Table~\ref{table5} show that incorporating both self distillation and attentive slices selection into our framework significantly improves the performance across all data splits and datasets, indicating the importance of key designs in \ourmethod{}.

\section{Conclusion}
In this paper, we propose a novel approach called \ourmethod{} that leverages diagnostic reports as a shared semantic space to create unified representations for diverse modalities of medical images, with a specific emphasis on 2D and 3D images. By using medical diagnostic reports as a bridge, we establish the unified vision-language framework that connects visual medical data across different modalities. Moreover, with the guidance of the text, we effectively extract visual modality information and accurately identify affected areas in 2D images and lesion slices in 3D CT scans, thereby enhancing consistency across various visual data modalities. Extensive experiments demonstrate \ourmethod{}'s superior performance in these downstream tasks(classification, segmentation, and retrieval) on various 2D and 3D medical image datasets. We hope our work can promote the exploration of VLP in medical image processing.

\bibliography{main}
\bibliographystyle{splncs04}

\newpage
\section*{Roadmap of Appendix}

The structure of the appendix is delineated as follows: Descriptions of the relevant experimental details are provided in the Section~\ref{sec1} and Section~\ref{sec2}. Subsequently, we conducted further experimental analysis in Section~\ref{sec3}. Lastly, Section~\ref{sec4} encompasses a presentation of supplementary visualization results.

\section{More Implementation Details of Fig.1(b)}
\label{sec1}
\subsection{Implementation Details}
We randomly compile an extensive collection of disease diagnostic reports from both 2D and 3D pre-training datasets, such as (1) 2D: ``concerning for pneumonia.'', ``Trace bilateral pleural effusions.'', ``Unchanged minimal left pleural effusion.'', ``Small left pleural effusion stable.'', (2) 3D: ``Mild right pleural effusion was also present.'', ``areas of opacity in frosted glass with peripheral predominance in both hemithoraces.'', ``both upper lobes and in the left lower lobe with a pattern of organizing pneumonia.'', ``Pulmonary edema present.''. Subsequently, we employ BioClinicalBERT~\cite{alsentzer2019publicly} to encode these diagnostic reports into corresponding tokens. Following this encoding process, we utilized t-SNE for visualization purposes, leading to the results depicted in Figure 1b. The outcomes illustrate that the representations of reports corresponding to 2D and 3D images coalesce within a shared semantic space.

\section{More Implementation Details of Downstream Tasks}
\label{sec2}
\subsection{Medical Classification} 
\noindent \textcolor{black}{\textbf{2D Medical Image Classification.} Except for the fine-tuning of the entire CheXpert dataset, where we use a batch size of 96, we use a batch size of 48 for the rest of the linear classification settings. Similar to the image preprocessing of MIMIC-CXR, we resize the larger dimension to 256 and pad zeros on the smaller side, resulting in an image size of 256 $\times$ 256. Then, we randomly crop (for training) or centered crop (for validation and testing) an image to 224 $\times$ 224 and normalize it into the range [0, 1] as the input for the model. The optimizer used is AdamW~\cite{loshchilov2017decoupled} with a learning rate of 5e-4 (except for COVIDx where we use 5e-3) and weight decay of 1e-6. We fine-tune the image classifier for 50 epochs and implement early stopping when the validation loss does not decrease for 10 consecutive runs. Afterward, we save the checkpoint model with the lowest validation loss for testing}.

\noindent \textcolor{black}{\textbf{3D Medical Image Classification.} (1) CC-CCII classification is to classify each volume into three categories: \emph{novel coronavirus pneumonia}, \emph{common pneumonia}, and \emph{normal}. We use a batch size of 8. We resize the 3D volumes to $32 \times 128 \times 128$. We use the randomflip to augment the train set. The optimier used is AdamW and we train the classifier for 50 epochs. (2) LUNA 16 \cite{setio2017validation}, which is established from LIDC-IDRI \cite{armato2011lung}. It finally contains 888 CT scans with annotations which removes CT scans with slice thickness greater than 3mm of LIDC-IDRI database. The task is a binary classification,  \emph{i.e.}, classifying each CT volume into \emph{pulmonary nodule} or \emph{normal}. The optimier used is AdamW and we train the whole network for 100 epochs}. \textcolor{black}{Our baseline methods include UniMiss~\cite{xie2022unimiss} and Joint~\cite{nguyen2023joint}, which belongs to 2D and 3D co-learning
methods. UniMiSS not only learns 2D and 3D representations but also concurrently learn all 2D sections derived from 3D volumes, along with all 2D X-ray data. Joint directly learn all 2D sections derived from 3D volumes, along with all 2D X-ray data through contrastive learning}.

\subsection{\textcolor{black}{Medical Segmentation.}}
\label{AppendixB.2}
\noindent \textcolor{black}{\textbf{2D Medical Image Segmentation.} In the case of the RSNA dataset, we create masks for the pneumonia-affected areas based on the provided bounding boxes. These images and corresponding masks are then resized to dimensions of 224 $\times$ 224. To augment the training set, we implement ShiftScaleRotate, encompassing random affine transformations such as translation, scaling, and rotation. Following this, the images are normalized to fall within the [0, 1] range before being supplied to the semantic segmentation model. we use the SETR-PUP (progressive upsample) architecture in~\cite{zheng2021rethinking} by replacing the encoder with \ourmethod{}. We freeze the pre-trained image encoder and only train decoder portion. The training process involves the use of the AdamW optimizer with a learning rate of 5e-4 and a weight decay of 1e-6. As suggested by~\cite{huang2021gloria}, we adopt a combined loss equation of $\alpha \times$ FocalLoss + DiceLoss, with $\alpha$ set to 10. The semantic segmentation model undergoes fine-tuning for 50 epochs, with batch size 16 and early stopping implemented if the validation loss ceases to decrease after 10 consecutive runs. The checkpoint model that exhibits the minimum validation loss is then preserved for testing}.

\noindent \textcolor{black}{\textbf{3D Medical Image Segmentation.}} \textcolor{black}{In the case of the BCV dataset, the images and correspoinding. The 3D volumes are resized to $48 \times 224 \times 224$. To augment the traning set, we implement random rotation, scaling, flipping, adding white Gaussian noise, Gaussian blurring, adjusting rightness and contrast, simulation of low resolution, and Gamma transformation. We use the UNETR~\cite{hatamizadeh2022unetr} architecture by replace the encoder with pre-trained \ourmethod{}. We train the entire model.. The training process involves the use of the AdamW optimizer with a learning rate of 1e-4. We adopt a combined loss equation of Dice + CE. The semantic segmentation model fintunes for 25,000 iterations with batch size 2.}

\section{\textcolor{black}{More Analysis}}
\label{sec3}
\subsection{\textcolor{black}{3D Medical Segmentation in AMOS}}
\textcolor{black}{AMOS is a large-scale, diverse, clinical dataset for abdominal organ segmentation, which is divided into 200/100 CTs for training/validation. The training details is the same as~\ref{AppendixB.2}. We report the Dice score (\%) training with 20\%, 40\%, and 100\% portion.}

\setlength{\tabcolsep}{4mm}
\begin{table}[t]
    \caption{\textcolor{black}{3D Semantic Segmentation results on AMOS (Dice [\%]). AMOS is fine-tuned with 20\%, 40\%, 100\% training data. Best results are in boldface.}}
    \centering
    \begin{tabular}{l c c c c c} 
        \hline
        {} & \multicolumn{3}{c}{\textcolor{black}{AMOS}} \\
        \textcolor{black}{Method} & \textcolor{black}{20\%} & \textcolor{black}{40\%} & \textcolor{black}{100\%} \\
        \midrule
        \textcolor{black}{UniMiss} & \textcolor{black}{\bf 79.5} & \textcolor{black}{82.3} & \textcolor{black}{85.8} \\
        \hline
        \rowcolor{graycolor}  \textcolor{black}{\ourmethod{}} & \textcolor{black}{78.8} & \textcolor{black}{{\bf 82.9}} & \textcolor{black}{{\bf 86.4}} \\ 
        \hline
    \end{tabular}
    \label{appendix_tab2}
\end{table}

\subsection{\textcolor{black}{Different Metrics in COVIDx}}

\setlength{\tabcolsep}{4mm}
\begin{table}[t]
    \centering
    \caption{\textcolor{black}{3D Semantic Segmentation results on AMOS (Dice [\%]). AMOS is fine-tuned with 20\%, 40\%, 100\% training data. Best results are in boldface.}}
    \begin{tabular}{l c c c c c c c c} 
        \hline
        {} & \multicolumn{3}{c}{\textcolor{black}{COVIDx (Acc/AUC)}}\\
        \textcolor{black}{Method} & \textcolor{black}{1\%} & \textcolor{black}{10\%} & \textcolor{black}{100\%} \\
        \midrule
        \textcolor{black}{MGCA} & \textcolor{black}{ 74.8/89.0} & \textcolor{black}{84.8/97.0} & \textcolor{black}{92.3/97.9} \\
        \hline
        \rowcolor{graycolor} \textcolor{black}{\ourmethod{}} & \textcolor{black}{\bf 80.3/93.5} & \textcolor{black}{{\bf 92.4/98.1}} & \textcolor{black}{{\bf 94.6/98.1}} \\ 
        \hline
    \end{tabular}
    \label{appendix_tab3}
\end{table}

\begin{figure}[htbp]
    \centering  
     \includegraphics[width=1 \textwidth]{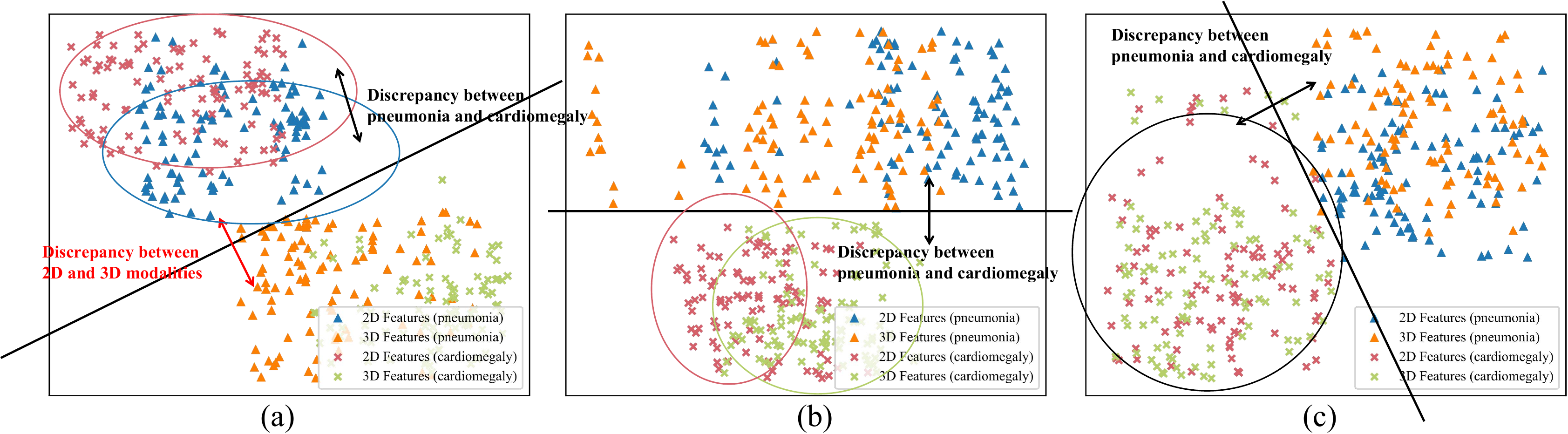}
    \caption{\textcolor{black}{t-SNE visualizations of image representations by models trained with different methods. (a) Two models for different image modalities are trained individually in separate VLP process. (b) One models for different image modalities are trained in one VLP processes, but without designes in \ourmethod{}. (c) \ourmethod{}. We use circles to highlight differences between different images.}}
\label{fig:appendix_tsne}
\end{figure} 

\textcolor{black}{We applied two distinct evaluation metrics, namely AUC (Area Under the Curve) and ACC (Accuracy), to assess the performance of our model on the COVIDx dataset. AUC is a widely used metric in machine learning and it represents the probability that a random positive example will be ranked higher than a random negative example. A higher AUC indicates better model performance. On the other hand, Accuracy (ACC) is a measure of how many predictions a model gets right out of all the predictions it makes. It is calculated as the number of correct predictions divided by the total number of predictions. The results of our evaluation using these metrics on the COVIDx dataset are presented in Table~\ref{appendix_tab3}. These findings provide insights into the robustness of our model.}

\section{\textcolor{black}{Visualization of Feature Quality}}
\label{sec4}

\textcolor{black}{We add three t-SNE visualization in Fig.~\ref{fig:appendix_tsne}. Compared to Fig.~\ref{fig:tsne}, Fig.~\ref{fig:appendix_tsne} add more class (cardiomegaly) to demonstrate the ability to unify different modal representations. We have marked the changes in distances between different modalities in the figure. As shown in Fig.~\ref{fig:appendix_tsne}, UniMedI effectively reduces the distance between different modalities and stablishes a unified representation space}.

\end{document}